\documentclass[10pt,twocolumn,letterpaper]{article}

\usepackage[pagenumbers]{cvpr} 

\usepackage[dvipsnames]{xcolor}

\usepackage{amsmath,amsfonts,bm}

\def\eqref#1{equation~\ref{#1}}

\def\1{\bm{1}}

\DeclareMathAlphabet{\mathsfit}{\encodingdefault}{\sfdefault}{m}{sl}
\SetMathAlphabet{\mathsfit}{bold}{\encodingdefault}{\sfdefault}{bx}{n}

\usepackage{graphicx}
\usepackage{amsmath}
\usepackage{amssymb}
\usepackage{booktabs}
\usepackage{enumerate}

\usepackage{amsfonts}
\usepackage{amsthm}
\usepackage{bbm}

\usepackage{wrapfig,lipsum,booktabs}
\usepackage{color}
\usepackage{algorithm,algpseudocode}
\usepackage{multicol}
\usepackage{array,multirow}

\usepackage{multirow}
\usepackage{tabularx}
\usepackage{listings}
\usepackage{xcolor}
\usepackage{color}
\usepackage{colortbl}

\makeatletter
\DeclareRobustCommand\onedot{\futurelet\@let@token\@onedot}
\def\@onedot{\ifx\@let@token.\else.\null\fi\xspace}

\makeatother
\definecolor{cvprblue}{rgb}{0.21,0.49,0.74}
\usepackage[pagebackref,breaklinks,colorlinks,citecolor=cvprblue]{hyperref}

\title{LightningDrag: Lightning Fast and Accurate \\Drag-based Image Editing Emerging from Videos}

\author{Yujun Shi\textsuperscript{\rm 1}$^*$
\quad\quad
Jun Hao Liew\textsuperscript{\rm 2}$^{*\dag}$
\quad\quad
Hanshu Yan\textsuperscript{\rm 2}
\quad\quad
Vincent Y. F. Tan\textsuperscript{\rm 1}
\quad\quad
Jiashi Feng\textsuperscript{\rm 2} \\
\textsuperscript{\rm 1}National University of Singapore
\quad\quad \textsuperscript{\rm 2} ByteDance Inc.\\
{\tt\small shi.yujun@u.nus.edu} \quad {\tt\small vtan@nus.edu.sg} \quad
{\tt\small jshfeng@bytedance.com}
}

\begin{document}

\twocolumn[{
\maketitle
\begin{center}
    \centering
    \captionsetup{type=figure}
    \vspace{-0.3cm}
    \includegraphics[width=\textwidth]{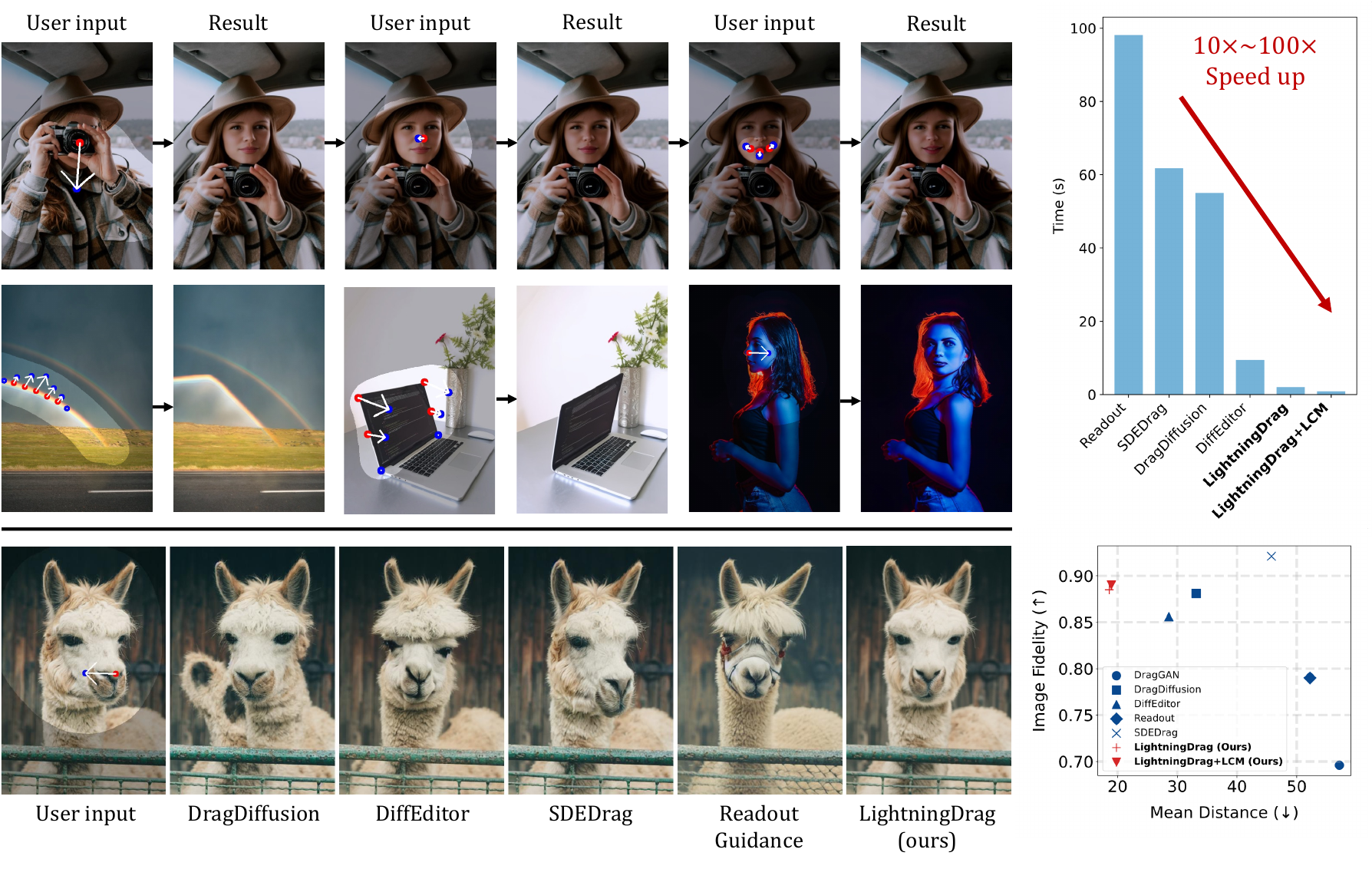}
    \caption{\textbf{{\sc LightningDrag} achieves high quality drag-based image editing under $1$ second.} The user provides handle points (\textcolor{red}{red}), target points (\textcolor{blue}{blue}), and a mask specifying the editable region (\textcolor{gray}{brighter area}). Lower Mean Distance indicates more effective ``draggging'', while higher Image Fidelity implies better appearance preserving. Our approach significantly surpass the previous methods in terms of both speed and quality. Image credit (source images): Pexels. Project page: \href{https://lightning-drag.github.io/}{https://lightning-drag.github.io/}.}
    \label{fig:teaser}
\end{center}
}]
{
  \renewcommand{\thefootnote}%
    {\fnsymbol{footnote}}
  \footnotetext[1]{These two authors make the equal contributions.}
  \footnotetext[2]{Project lead.}
}

\begin{abstract}
  Accuracy and speed are critical in image editing tasks. Pan et al.\ introduced a drag-based image editing framework that achieves pixel-level control using Generative Adversarial Networks (GANs). A flurry of subsequent studies enhanced this framework's generality by leveraging large-scale diffusion models. However, these methods often suffer from inordinately long processing times (exceeding $1$ minute per edit) and low success rates. Addressing these issues head on, we present {\sc LightningDrag}, a rapid approach enabling high quality drag-based image editing in $\sim1$ second. Unlike most previous methods, we redefine drag-based editing as a conditional generation task, eliminating the need for time-consuming latent optimization or gradient-based guidance during inference. In addition, the design of our pipeline allows us to train our model on large-scale paired video frames, which contain rich motion information such as object translations, changing poses and orientations, zooming in and out, \emph{etc}. By learning from videos, our approach can significantly outperform previous methods in terms of accuracy and consistency. Despite being trained solely on videos, our model generalizes well to perform local shape deformations not presented in the training data (e.g., lengthening of hair, twisting rainbows, etc.). Extensive qualitative and quantitative evaluations on benchmark datasets corroborate the superiority of our approach. The code and model will be released at \href{https://github.com/magic-research/LightningDrag}{https://github.com/magic-research/LightningDrag}.
\end{abstract}

\maketitle

\section{Introduction}
Image editing using generative models~\cite{roich2022pivotal,endo2022user,hertz2022prompt,mokady2023null,kawar2023imagic,parmar2023zero} has received considerable attention in recent years. However, many existing approaches lack the ability to conduct fine-grained spatial control. One landmark work attempting to achieve precise spatial image editing is {\sc DragGAN}~\cite{pan2023drag}, which enables interactive point-based image manipulation on generative adversarial networks (GANs). Using their method, users initiate the editing process by selecting pairs of handle and target points on an image. Subsequently, the model executes semantically coherent edits by relocating the contents of the handle points to their corresponding targets. Moreover, users have the option to delineate editable regions using masks, preserving the integrity of the rest of the image. Building upon the foundation laid by \citet{pan2023drag}, subsequent works \cite{shi2023dragdiffusion,mou2023dragondiffusion,nie2023blessing,ling2023freedrag} have endeavored to extend this editing framework to large-scale pre-trained diffusion models \cite{rombach2022high}, aiming to further enhance its generality.

However, a common drawback among many methods within this framework is their lack of efficiency.
Prior to editing a real image input by the user, {\sc DragGAN} \cite{pan2023drag} requires applying a lengthy pivotal-tuning-inversion \cite{roich2022pivotal}, a process that can consume up to $1$ to $2$ minutes. As for diffusion-based approaches such as {\sc DragDiffusion} \cite{shi2023dragdiffusion} and {\sc DragonDiffusion} \cite{mou2023dragondiffusion}, they typically entail time-consuming operations such as latent-optimization or gradient-based guidance during editing.
This inefficiency poses a significant barrier to practical deployment in real-world scenarios. What undermines the users' experiences even more is the low success rate of these methods. Since they are mostly zero-shot methods that lack explicit supervision to perform drag-based editing, they frequently struggle with either accurately moving semantic content from handle to target points or preserving the appearance and identity of the source image.

In this study, we introduce {\sc LightningDrag}, a novel approach that achieves state-of-the-art drag-based editing while drastically reducing latency to less than $1$ second, thereby making drag-based editing {\em highly practical} for deployment. To attain such rapid drag-based editing, we redefine the task as a specific form of conditional generation, where the source image and the user's drag instruction serve as conditions. Drawing inspiration from previous literature \cite{ref_only_controlnet,xu2023magicanimate,hu2023animate,chen2024wear,alzayer2024magic}, we leverage the reference-only architecture to process source images for identity preservation. Additionally, to incorporate the user's drag instruction into the generation process, we encode the handle and target points into corresponding embeddings via a Point Embedding Network. These embeddings are then injected into self-attention modules of the backbone diffusion model to guide the generation process.
This approach eliminates the need for repeatedly computing gradients on diffusion latents during inference, as had been done in previous methods, thereby significantly reducing latency to that of generating an image with diffusion models. 
As a conditional generation pipeline, our approach can be further accelerated by integrating off-the-shelf acceleration modules for diffusion models (\emph{e.g.}, LCM-Lora \cite{luo2023lcm}, PeRFlow \cite{yan2024perflow}), a capability not possible with previous gradient-based methods.

\begin{figure*}
    \centering
    \includegraphics[width=\textwidth]{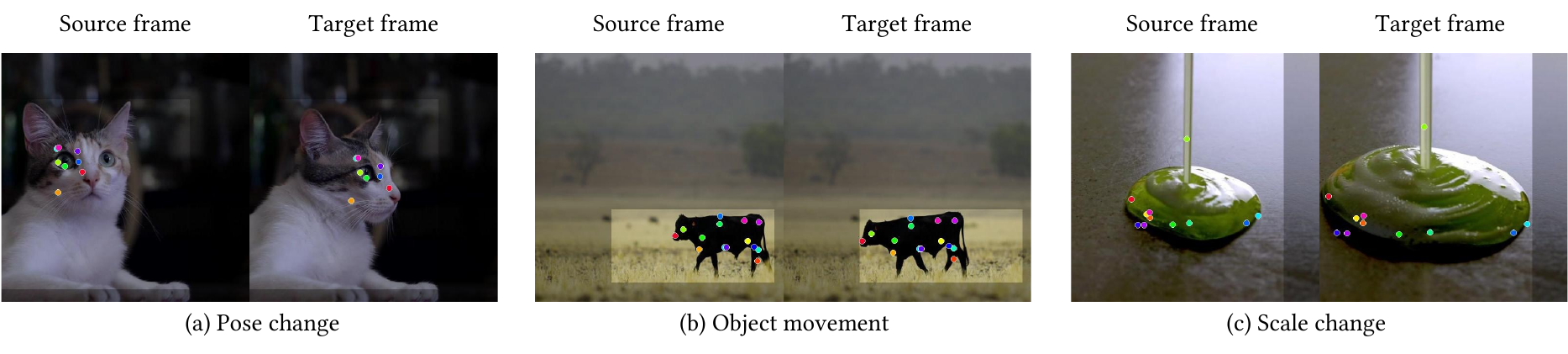}
    \caption{\textbf{Samples of collected supervision pairs from videos.} Video motion contains various transformation cues such as pose change, object movement and scale change, which are useful for the model to learn how objects change and deform while avoiding appearance change.}
    \label{fig:video_data_pipeline}
\end{figure*}

To train our proposed model, we leverage video frames as our supervision signals. This choice is motivated by the fact that video motions inherently encapsulate transformations relevant to drag-based editing (Fig. \ref{fig:video_data_pipeline}), such as object translations, changing poses and orientations, zooming in and out, \emph{etc}.
Our training data is constructed from paired video frames.
Firstly, we sample pixels that exhibit significant optical flow magnitude on the first frame as the handle points. Next, we employ CoTracker2 \cite{karaev2023cotracker} to identify the handle points' corresponding target points in the second frame. This procedure allows us to construct training pairs for our model on a large scale. By learning from such large-scale video frames, our approach significantly outperforms previous methods in terms of both accuracy and consistency.
One potential concern regarding our data construction pipeline is that certain transformations involving local deformation (\emph{e.g.}, lengthening of hair, twisting rainbows, \emph{etc.}) are not explicitly presented in video motions. However, intriguingly, we find that our model  generalizes well to these out-of-domain editing instructions after being trained on videos.

Through comprehensive evaluation across a wide array of samples, encompassing images of diverse categories and styles, we showcase the substantial advantages of our approach in terms of both speed and quality. As illustrated in Fig.~\ref{fig:teaser}, our approach adeptly delivers editing results in accordance to the user's instructions with an imperceptible latency of less than $1$ second. Furthermore, we delve into two key techniques, namely \textit{source noise prior} and \textit{point-following classifier-free guidance}, that enhance the accuracy and consistency of our pipeline during inference.
Lastly, we explore two test-time strategies that users can employ to further refine drag-based editing results---\textit{point augmentation} and \textit{sequential dragging}.

\section{Related Works}
\textbf{Generative image editing.} In light of the initial successes achieved by generative adversarial networks (GANs) in image generation \citep{NIPS2014_5ca3e9b1,karras2019style,karras2020analyzing}, a plethora of image editing techniques have emerged based on the GAN framework \citep{endo2022user,pan2023drag,abdal2021styleflow,leimkuhler2021freestylegan,patashnik2021styleclip,shen2020interpreting,shen2021closed,tewari2020stylerig,harkonen2020ganspace,zhu2016generative,zhu2023linkgan}. However, owing to the limited model capacity of GANs and the inherent challenges in inverting real images into GAN latents \citep{abdal2019image2stylegan,creswell2018inverting,lipton2017precise,roich2022pivotal}, the applicability of these methods is inevitably restricted. Recent advancements in large-scale text-to-image diffusion models \citep{rombach2022high,saharia2022photorealistic} have spurred a surge of diffusion-based image editing methods \citep{hertz2022prompt,cao2023masactrl,mao2023guided,kawar2023imagic,parmar2023zero,liew2022magicmix,mou2023dragondiffusion,tumanyan2023plug,brooks2023instructpix2pix,meng2021sdedit,bar2022text2live,epstein2023selfguidance}. While many of these methods aim to manipulate images using textual prompts, conveying editing instructions through text presents its own set of challenges. Specifically, the prompt-based paradigms are often limited to alterations in high-level semantics or styles, lacking the precise spatial control.

\noindent \textbf{Point-based image editing.} Point-based image editing is a challenging task aiming to manipulate images in pixel-level precision. Traditional literature in this field \cite{beier2023feature,igarashi2005rigid,schaefer2006image} have relied on non-parametric techniques. However, recent advancements driven by deep learning-based generative models, such as GANs, have propelled this field forward, with several notable contributions \cite{pan2023drag,endo2022user,wang2022rewriting,zhu2016generative}.
One notable work among these is \citet{pan2023drag}, which achieves impressive interactive point-based editing by optimizing GAN latent codes.
Nonetheless, the applicability of this framework is limited by the inherent capacity constraints of GANs. In a bid to enhance its versatility, subsequent efforts have endeavored to extend the framework to large-scale diffusion models \cite{shi2023dragdiffusion,mou2023dragondiffusion,luo2023readout,geng2024motion,cui2024stabledrag,liu2024drag}. However, most of these works still rely on computationally intensive operations such as latent optimization or gradient-based guidance, necessitating repeated gradient computations on diffusion latents and rendering them impractical for real-world deployment. Different from these works, \citet{nie2023blessing} introduce a paradigm that obviates the need for gradient computation on diffusion latents. However, this paradigm still requires repeated diffusion-denoising operations, resulting in latencies comparable to gradient-based methods such as \citet{shi2023dragdiffusion}. Recent works by \citet{li2024dragapart} and \citet{chen2024wear} redefine the drag-based editing into a generation task, drastically reducing the editing latency to levels comparable to generating images with diffusion models. However, these studies are narrowly focused: \citet{li2024dragapart} delve into modeling part-level movements in articulated objects, while \citet{chen2024wear} concentrates on single human images with clothes. In contrast, our approach enables lightning-fast drag-based editing on \emph{general images}.

\noindent \textbf{Learning image editing from videos}
Previous methods leveraging videos to aid in learning image editing typically sample two frames from a video to form a supervision pair. For instance, \citet{chen2023anydoor} utilize collected image pairs for the same object from videos to learn the appearance variations, thus improving their subject-composition pipeline. \citet{alzayer2024magic} use video frames to supervise their proposed coarse-to-fine warping-based image editing pipeline. \citet{luo2023readout} propose pre-training diffusion models on video data to enhance drag-based editing performance. However, their approach still relies on time-consuming gradient-based guidance and is trained on a limited dataset comprising only around $100$ supervision pairs. In contrast to these works, we train a conditional generation pipeline on \emph{large-scale} video data to perform fast and accurate drag-based editing.

\section{Preliminaries}
\subsection{Latent Diffusion Models}
Diffusion models \cite{sohl2015deep,ho2020denoising} demonstrate promising performance in visual synthesis. \citet{rombach2022high} proposed the latent diffusion model (LDM), which first maps a given image $x_0$ into a lower-dimensional space via a variational auto-encoder (VAE) \citep{kingma2013auto} to produce $z_0 = \mathcal{E}(x_0)$. Then, a diffusion model with parameters $\theta$ is used to approximate the distribution of $q(z_0)$ as the marginal $p_{\theta}(z_0)$ of the joint distribution between $z_{0}$ and a collection of latent random variables $z_{1:T} = (z_1, \ldots, z_T)$. Specifically,
\begin{equation}
    p_{\theta}(z_{0}) = \int{p_{\theta}(z_{0:T})\, \mathrm{d}z_{1:T}},
\end{equation}
where $p_{\theta}(z_T)$ is a standard normal distribution and the transition kernels $p_{\theta}(z_{t-1}|z_t)$ of this Markov chain are all Gaussian conditioned on $z_t$. In our context, $z_{0}$ corresponds to the VAE latent of image samples given by users, $z_{t}$ is the latent after $t$ steps of the diffusion process. Specifically, 
\begin{equation}
\label{eq:add_noise}
    z_t = \sqrt{\bar{\alpha}_t}z_0 + \sqrt{1-\bar{\alpha}_t}\epsilon,
\end{equation}
where $\epsilon \sim \mathcal{N}(0,\mathbf{I})$, and $\bar{\alpha_t}$ is the cumulative product of the noise coefficient $\alpha_t$ at each step.

Based on the framework of LDM, several powerful pretrained diffusion models have been released publicly, including the Stable Diffusion (SD) model (\href{https://huggingface.co/stabilityai}{https://huggingface.co/stabilityai}). In this work, our proposed pipeline is developed based on SD model.


\section{Methodology}
In this section, we formally present our {\sc LightningDrag} approach. To start, we elaborate on the details of how we construct supervision pairs for our model from videos in Sec.~\ref{sec:data_collection}. Next, we describe the architecture design of our model in Sec.~\ref{sec: arch_design}. Furthermore, we introduce some techniques we use during test-time to improve the editing results in Sec.~\ref{sec:test_time_tech}. Finally, we introduce some strategies that users can employ to fix failure cases in Sec.~\ref{sec:test_time_strategies}.

\subsection{Paired supervision from video data}
\label{sec:data_collection}
One of the challenges we encounter is in collecting large-scale paired data for training the model, as obtaining user-annotated input-output pair on a large scale is nearly infeasible. In this work, we redirect our focus towards leveraging video data. Our key insight lies in the inherent motion captured within video, which naturally encompasses various transformations relevant to drag-based editing, including zooming in and out, changes in pose and orientation, \emph{etc.} These dynamics offer valuable cues for the model to learn how objects undergo changes and deform.

We begin by curating videos with static camera movement, simulating drag-based editing where only local regions are manipulated while others remain static. Subsequently, we randomly sample two frames from these videos to serve as source $I_{\rm src}$ and target images $I_{\rm tgt}$, respectively. We will resample another pair if the optical flow between the two images is too small. Next, we sample $N$ handle points $P_{\rm hdl}$ on $I_{\rm src}$ with a probability proportional to the optical flow strength, ensuring the selection of points with significant movement. We then employ CoTracker2 \cite{karaev2023cotracker}, a state-of-the-art point tracking algorithm to extract the corresponding target points $P_{\rm tgt}$ in the target image $I_{\rm tgt}$. Finally, we adopt a similar approach as in~\citet{dai2023animateanything} to extract a binary mask $M$ highlighting the motion areas, indicating regions to be edited. Collectively, the tuple $(I_{\rm src}, I_{\rm tgt}, P_{\rm hdl}, P_{\rm tgt}, M)$ form our training samples to train our {\sc LightningDrag}. Examples showcasing the versatility of video data for training drag-based editing can be found in Fig.~\ref{fig:video_data_pipeline}.

\subsection{Architecture Design}
\label{sec: arch_design}
\begin{figure}
    \centering
    \includegraphics[width=0.47\textwidth]{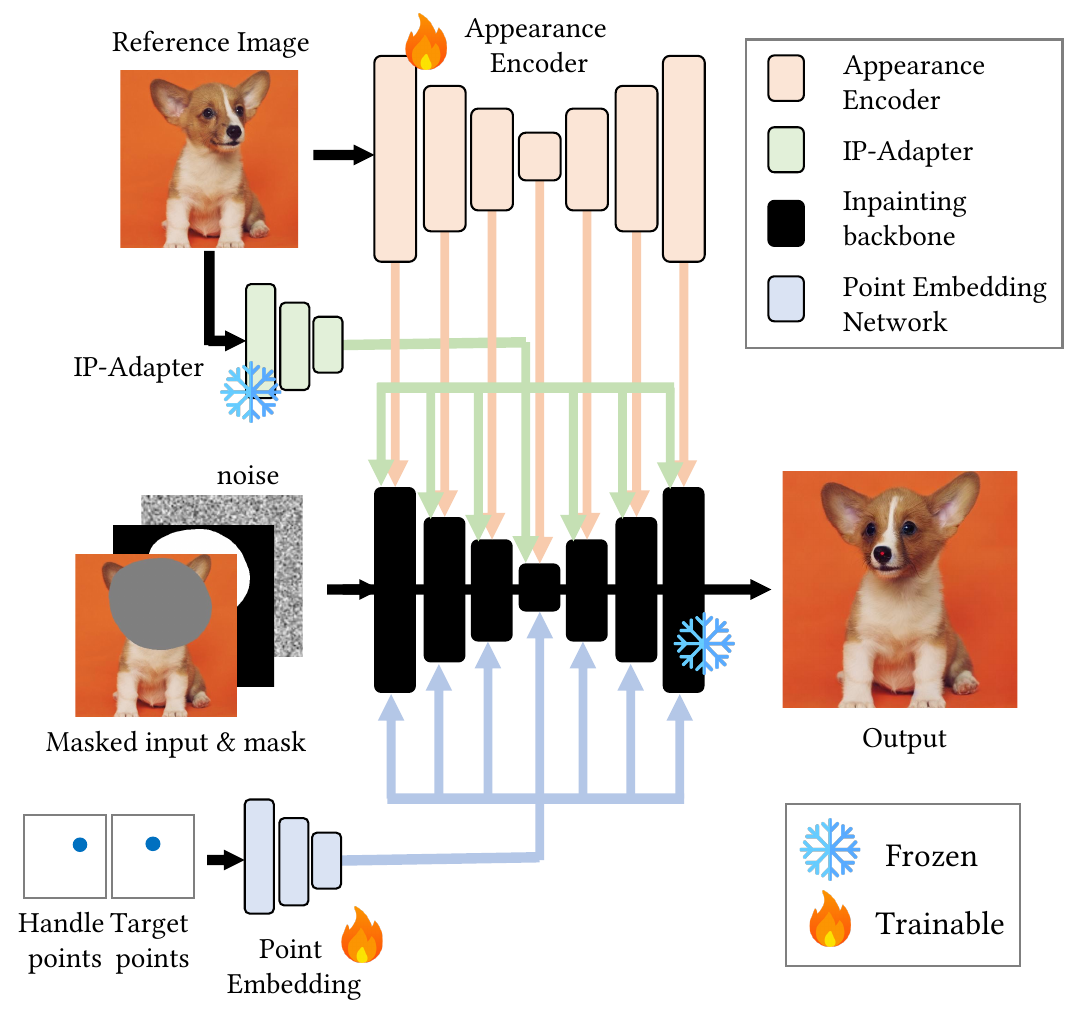}
    \caption{\textbf{The pipeline of {\sc LightningDrag}.} Our {\sc LightningDrag} consists of three components, including (1) an inpainting diffusion backbone to enforce unmasked regions remain untouched; (2) an Appearance Encoder for preserving the identity of the reference image; and (3) a Point Embedding Network to encode the (handle, target) points pairs. }
    \label{fig:pipeline}
\end{figure}

We formulate the drag-based image editing task as a conditional generation problem, where the generated image needs to fulfill the following criteria: (1) unmasked area remains untouched; (2) image identity (\emph{e.g.}, human face, texture, \emph{etc.}) should be preserved after dragging; (3) the areas indicated by handle points should move to the target coordinates. To achieve this, our {\sc LightningDrag} comprises three components: (1) an image inpainting backbone to enforce unmasked region remains identical; (2) an appearance encoder preserves identity of $I_{\rm src}$, (3) a point embedding network encodes the (handle, target) points pairs, accompanied by a point-following attention mechanism, which explicitly enables the model to follow the point instructions.
The overall framework is depicted in Fig.~\ref{fig:pipeline}. We next elaborate on each component in more details.

\subsubsection{Inpainting Backbone.} We utilize the Stable Diffusion Inpainting U-Net as our backbone, which takes concatenation of the following as input: noise latents $z_t$, a binary mask $M$, and masked latents $M \odot z_0^{\rm src}$. It is worth noting that the inpainting backbone typically takes in a text prompt to indicate the inpainted content. However, in drag-based editing application, a text prompt is not only redundant as the image content is already provided by the source image, but also difficult for users to provide. Instead, we extract the image feature of the source image using IP-Adapter \cite{ye2023ip} and use an empty text prompt, freeing the users from this requirement.

\subsubsection{Appearance Encoder.}
To maintain the identity of the reference image, we draw inspiration from recent works on ID-consistent generation, such as \citet{xu2023magicanimate,hu2023animate,chen2024wear}. Specifically, we employ the reference-only architecture \cite{ref_only_controlnet} to process the source image. Unlike CLIP image encoder \cite{radford2021learning} which can only guarantee the overall colors and semantics, the reference-only approach has demonstrated efficacy in preserving fine-grained details of the reference image. Inherited from the weights of a pre-trained text-to-image U-Net diffusion model, our Appearance Encoder takes the reference latents $z_0^{\rm src}$ as input. It extracts the reference feature maps from the self-attention layers, which are subsequently used to guide the self-attention process in the denoising backbone.
The self-attention in the backbone is thus defined as follows:
\begin{equation}
\begin{aligned}
\label{eq:appr}
    \mathrm{Attn}(Q, K, V, K_{\mathrm{ref}}, V_{\mathrm{ref}})=\mathrm{Softmax} \Big(\frac{Q[K, K_{\mathrm{ref}}]^\top}{\sqrt{d}} \Big) [V, V_{\mathrm{ref}}],
\end{aligned}
\end{equation}
where $K_{\mathrm{ref}}$ and $V_{\mathrm{ref}}$ denote the keys and values extracted from the reference features, and $[\cdot,\cdot]$ denotes the concatenation operator.
Following prior works \cite{xu2023magicanimate}, we use clean reference latents as inputs to the Appearance Encoder (as opposed to noised latents used in original reference-only model \cite{ref_only_controlnet}). As a result, unlike backbone UNet that requires multiple denoising steps, the Apppearance Encoder only needs to extract features once throughout the entire editing process, which improves the model inference efficiency.

\subsubsection{Point Embedding Attention.}
Given the user-specified handle and target points, we first convert them into a handle and a target point map that is of the same resolution of the input image. Specifically, we randomly assign each pair of handle and target points with an integer number $k\in \{1,2,\ldots,N\}$, where $N$ is the maximum allowed points. Then, we put the integer $k$ to the pixel location on the point map given coordinates specified by handle and target points. The rest of the pixel locations on handle and target point maps are with value~$0$.

Once obtaining the handle and target point maps, we encode them into embedding via a point embedding network, which is composed of $12$ layers of convolution and SiLU activation. This network outputs embedding at four different resolutions, corresponding to the four different resolutions of SD UNet activation maps. To enable the model to follow point instructions effectively, we draw inspiration from \citet{chen2024wear} and introduce a point-following mechanism into Eqn. \ref{eq:appr}, resulting in the following formulation:
\begin{equation}
\begin{aligned}
\label{eq:appr_point}
    &\mathrm{Attn}(Q, K, V, K_{\mathrm{ref}}, V_{\mathrm{ref}}, E_{\mathrm{hdl}}, E_{\mathrm{tgt}}) \\ &=\mathrm{Softmax} \Big(\frac{(Q+E_{\mathrm{tgt}})[K+E_{\mathrm{tgt}}, K_{\mathrm{ref}}+E_{\mathrm{hdl}}]^\top}{\sqrt{d}} \Big) [V, V_{\mathrm{ref}}]
\end{aligned}
\end{equation}
where $E_{\mathrm{hdl}}$ and $E_{\mathrm{tgt}}$ are embeddings of handle and target point maps, respectively. In this way, we explicitly strengthen the similarity between the target points of the generated images and the handle points of the user input image, facilitating learning of drag-based editing.

\subsection{Test-time Techniques to Improve Editing Results}
\label{sec:test_time_tech}
\subsubsection{Noise prior}
\begin{figure}
    \centering
    \includegraphics[width=0.47\textwidth]{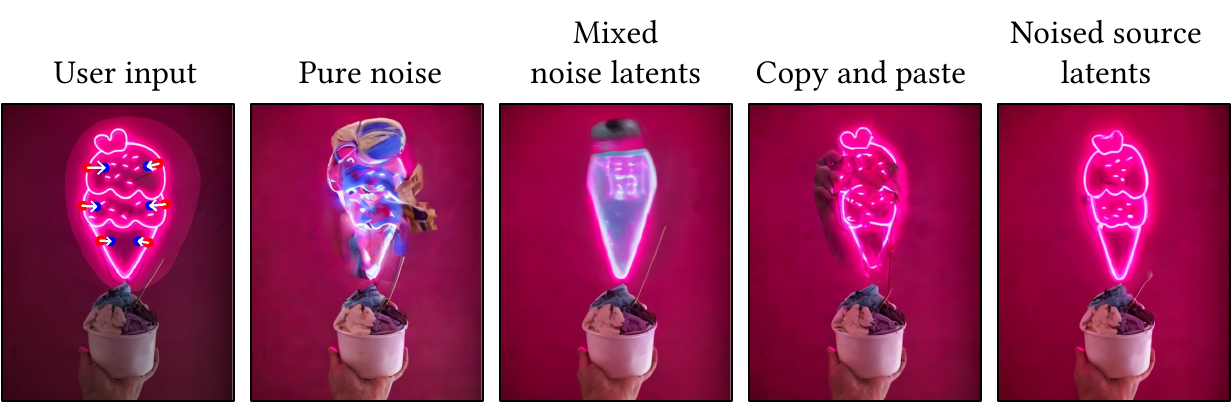}
    \caption{\textbf{Different strategies for constructing the noise prior.} We find that the ``noise source latents'' strategy produces the best results. Image credit (source image): Pexels}
    \label{fig:noise_prior}
\end{figure}

We have observed that directly using randomly initialized noise latents for generation sometimes yields unstable results, as depicted in Fig.~\ref{fig:noise_prior}. This instability may stem from the discrepancy between the initial noise during training and testing of diffusion models, as discussed in prior works \cite{lin2024common,lin2024sdxl}. In contrast to text-to-image generation, where obtaining a suitable initial noise prior is challenging, our task allows for a more accurate initialization of the noise prior by adding noise to the VAE latent of the source image. This technique enables us to narrow the gap between training and testing, resulting in more stable outcomes.

We ablate on the following strategies to construct the noise prior:
\begin{itemize}
    \item \textit{Noised source latents.} This strategy directly add noise on the source image latent with Eqn.~\eqref{eq:add_noise} to the terminal diffusion time-step of $t=999$.
    \item \textit{Mixed noise latents.} Based on the noised source latents, we re-initialize the user-provided mask region of the noise latents with pure Gaussian noise for potentially better editing flexibility.
    \item \textit{Copy and paste noise latents.} We borrow the ``copy and paste'' strategy from \citet{nie2023blessing} and apply it along with the handle and target points to obtain the initial noise prior.
\end{itemize}

We compare these noise prior strategies along with directly using pure random noise. We find the noised source latents produces the best results among these strategies, which is adopted to construct noise prior for our pipeline.

\subsubsection{Point-following classifier-free guidance}
\label{sec:point_follow_cfg}
To further improve the model's capability to follow the point instruction during inference, we implement the following \textit{point-following classifier-free guidance} (PF-CFG) to strengthen the effects of given (handle, target) points pairs:
\begin{equation}
\begin{aligned}
    \tilde{\epsilon}_{\theta}(z_t,c_{\mathrm{appr}},&c_{\mathrm{points}}) = \epsilon_{\theta}(z_t,c_{\mathrm{appr}}, \emptyset) \\
    &+ \omega(t)  \big(\epsilon_{\theta}(z_t,c_{\mathrm{appr}}, c_{\mathrm{points}}) - \epsilon_{\theta}(z_t,c_{\mathrm{appr}}, \emptyset) \big),
\end{aligned}
\end{equation}
where $\omega(t)$ is the time-dependent CFG scale, $c_{\mathrm{appr}}$ denotes the source image condition encoded by appearance encoder, and $c_{\mathrm{points}}$ denotes the condition of handle and target points.
To be more specific, when computing $\epsilon_{\theta}(z_t,c_{\mathrm{appr}}, \emptyset)$, we use Eqn.~\eqref{eq:appr} in all self-attention layers of the main backbone UNet. When computing $\epsilon_{\theta}(z_t,c_{\mathrm{appr}}, c_{\mathrm{points}})$, we employ Eqn.~\eqref{eq:appr_point}.

Most previous works involving diffusion models apply a fixed CFG scale across different denoising time-steps. However, recent literature \cite{kynkaanniemi2024applying,wang2024analysis} demonstrate the benefits of using a time-dependent CFG scale during denoising. In this work, we similarly find that a dynamic time-dependent CFG scale can help strike an approrpriate balance between the accuracy of point-following and image quality of the results.

\begin{figure}
    \centering
    \includegraphics[width=0.47\textwidth]{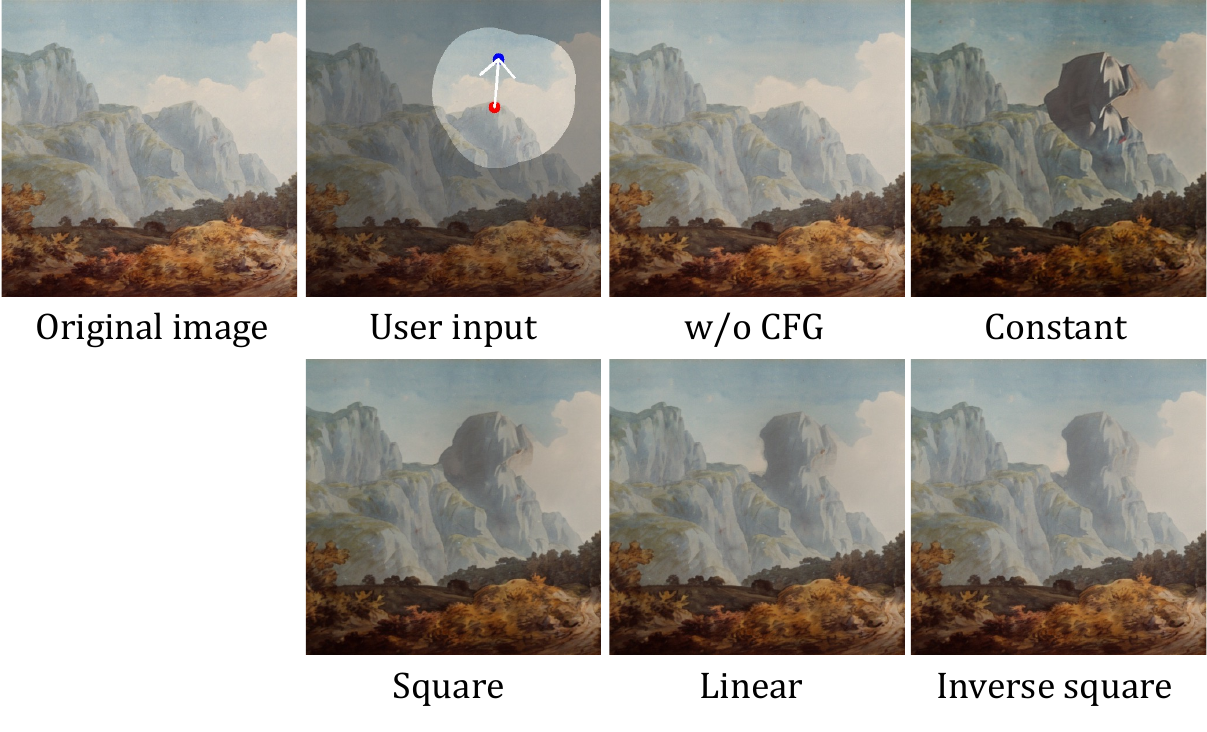}
    \caption{\textbf{Effects of different CFG scale schedules.} Our model struggles to conduct a successful drag when CFG is not used. Constant CFG scale often leads to over-saturation problem. On overall, fast decaying strategy (\textit{Inverse square}) attains the best results.}
    \label{fig:cfg_scheduler_qualitative}
\end{figure}

Denoting $\omega_{\mathrm{max}}$ as the maximum value of CFG, we explore the following CFG scale schedules: 
\begin{itemize}
    \item \textit{No CFG:} $\omega(t) = 1$
    \item \textit{Constant:} $\omega(t) = \omega_{\mathrm{max}}$.
    \item \textit{Square:} $\omega(t) = \omega_{\mathrm{max}}\times (1-(1-t/1000)^2) + (1-t/1000)^2$.
    \item \textit{Linear:} $\omega(t) = \omega_{\mathrm{max}}\times t/1000 + (1 - t/1000)$.
    \item\textit{Inverse square:} $\omega(t) = (\omega_{\mathrm{max}}-1)\times (t/1000)^2 + 1$.
\end{itemize}
We compare these schedules in Fig.~\ref{fig:cfg_scheduler_qualitative}. As can be observed, without using our CFG, the model struggles to conduct successful drag-based editing. On the other hand, using CFG with a constant scale can successfully drag the handle points to the target, but the results may suffer from over-saturation. By using schedules that decay the CFG scale from $\omega_{\mathrm{max}}$ to $1.0$ during the denoising process such as \textit{Square}, \textit{Linear}, and \textit{Inverse square}, we achieve accurate drag-based editing while markedly improve the image quality. Among these decaying schedules, we find fast decaying strategy such as \textit{Inverse square} achieves the best image quality, while slow decaying strategy such as \textit{Linear} and \text{Square} still suffer from slight quality degradation (\emph{e.g.,} over-saturation) on generated images.

\subsection{``Drag engineering'' to improve the editing}
Inspired by the use of prompt engineering technique in Large Language Models (LLM) to obatin ideal answers, we find that some failure cases produced by our {\sc LightningDrag} can also be mitigated by engineering the input drag instruction. Here, we introduce two strategies, namely \textit{Point augmentation} and \textit{Sequential dragging}, for users to consider when facing imperfect results with our model.

\label{sec:test_time_strategies}
\subsubsection{Point augmentation}
\begin{figure}
    \centering
    \includegraphics[width=0.47\textwidth]{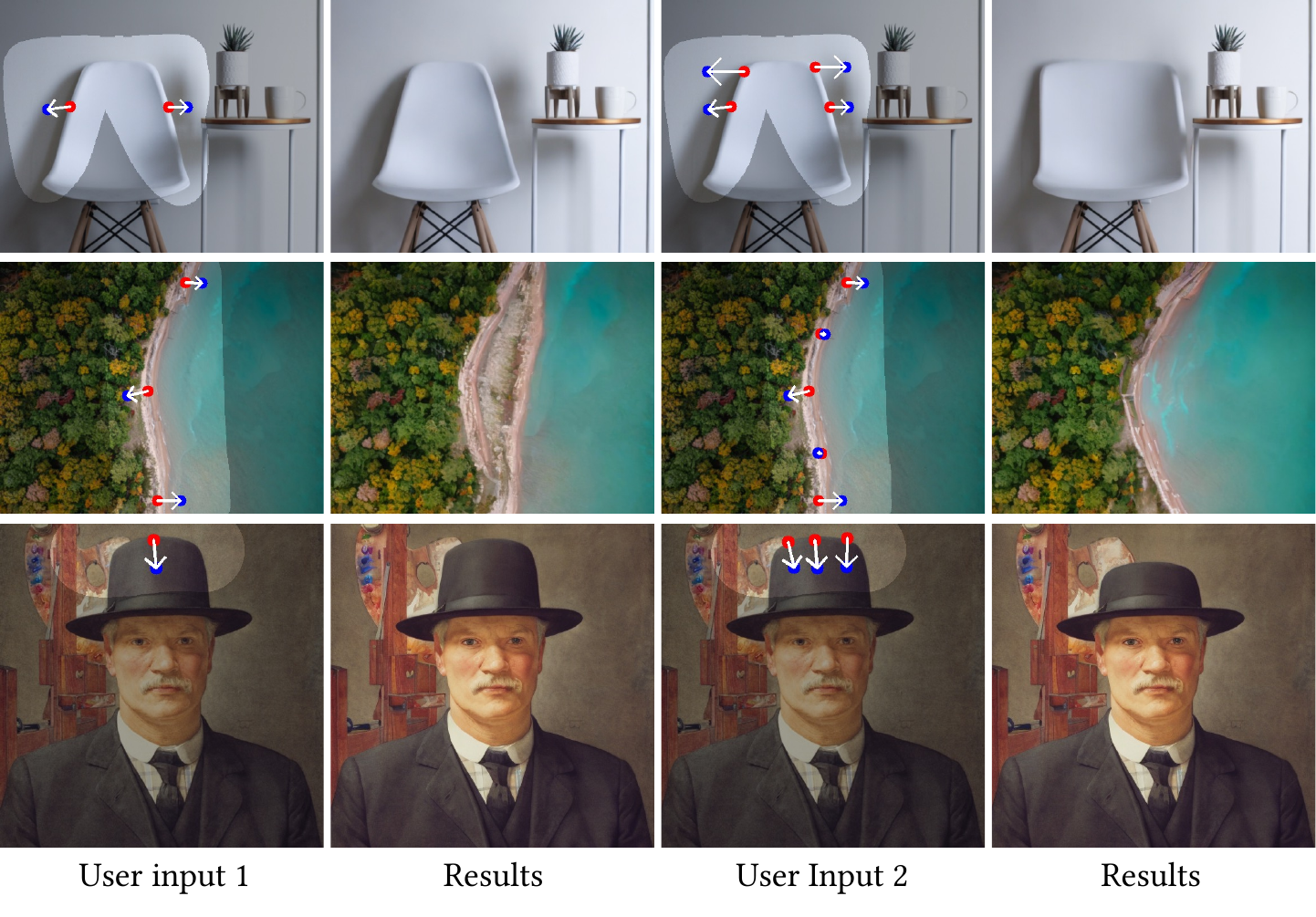}
    \caption{\textbf{Point Augmentation.} Augmenting with additional pairs of handle and target points can better convey the user's editing intention, which often leads to better performance.}
    \label{fig:point_aug}
\end{figure}

When the region specified by handle points fail to move to the target locations, augmenting the drag instruction with additional pairs of handle and target points has proven effective in improving results. Examples showcasing this augmentation are depicted in Fig.~\ref{fig:point_aug}. It is evident that by incorporating more pairs of handle and target points, users' editing intentions can be more explicitly conveyed, resulting in better outcomes. 

\subsubsection{Sequential dragging}
\begin{figure*}
    \centering
    \includegraphics[width=\textwidth]{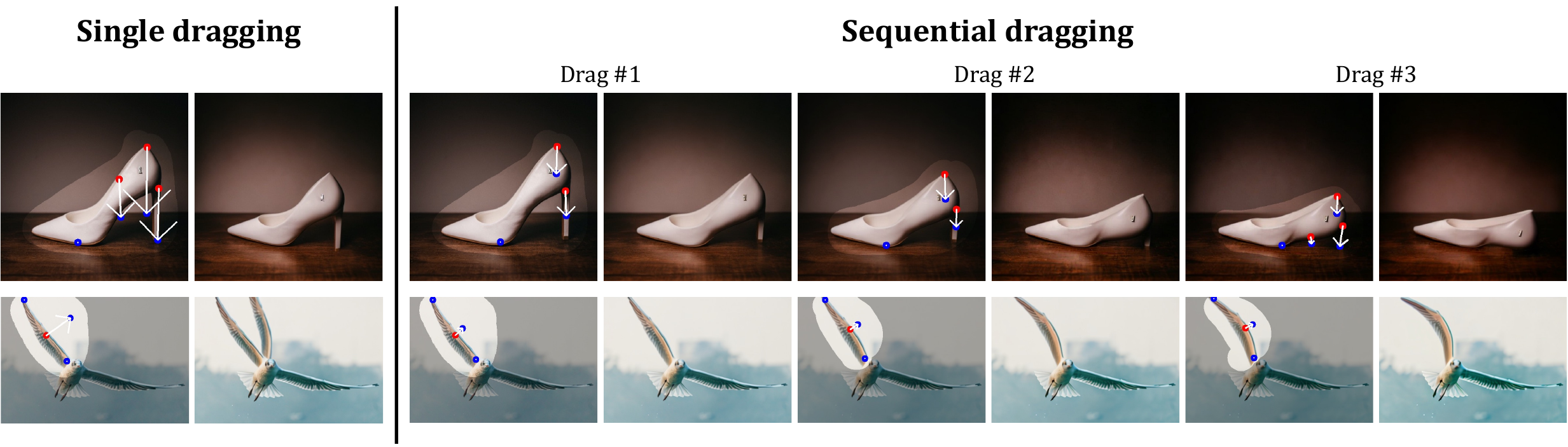}
    \caption{\textbf{Sequential dragging.} In cases when single dragging operation cannot attain  the desired outcome, a simple workaround is to break the operation down  into a sequence of shorter dragging trajectories. Image credit (source images): Pexels.}
    \label{fig:sequential_drag}
\end{figure*}

In cases where drag editing results are sub-optimal after one round of editing, users may opt to break down the drag instruction into multiple rounds and sequentially move semantic contents from handle points to final targets. Examples illustrating how such sequential dragging can rectify certain failure cases are presented in Fig.~\ref{fig:sequential_drag}. This strategy is facilitated by our model's exceptional ability to maintain the appearance and identity of the source image during editing. Without this capability, cumulative appearance shifts might occur, leading to undesired results. Additionally, given our model's negligible latency, employing sequential dragging does not significantly undermine user experience.

\section{Experiments}

\subsection{Implementation details}
\noindent \textbf{Network.}
The base inpainting U-Net inherits the pre-trained weights from Stable Diffusion V1.5 inpainting model\footnote{\href{https://huggingface.co/runwayml/stable-diffusion-inpainting}{https://huggingface.co/runwayml/stable-diffusion-inpainting}}, whereas the Appearance Encoder is initialized from the pre-trained weights of Stable Diffusion V1.5. The Point Embedding Network is randomly initialized, except for the last convolution layer which is zero-initialized \cite{zhang2023adding} to ensure the model starts training as if no modification has been made.

\smallskip \noindent \textbf{Training.}
We sample $220\mathrm{k}$ training samples from our internal video dataset to train our model. We set the learning rate to $5e-5$ with a batch size of $256$. We freeze both the inpainting U-Net and IP-Adapter, training both Appearance Encoder and Point Embedding Network. During training, we randomly sample $[1, 20]$ points pairs. We randomly crop a square patch covering the sampled points and resize to $512\times512$. 

\smallskip \noindent \textbf{Inference.}
We use DDIM \cite{song2020denoising} sampling with $25$ steps for inference by default. We found that our model is also compatible with recent diffusion acceleration techniques such as LCM-LoRA \cite{luo2023lcm} and PeRFlow \cite{yan2024perflow} without additional training. When using LCM-LoRA or PeRFlow, we use $8$ steps for sampling. We use guidance scale $\omega_{\mathrm{max}}$ of $3.0$ and adopt an inverse square decay (Sec.~\ref{sec:point_follow_cfg}) that gradually reduces the guidance scale to $1.0$ over time to prevent over-saturation issue.

\subsection{Evaluation on DragBench}
\label{drag_evaluation}
We provide a quantitative assessment of  our method on DragBench \cite{shi2023dragdiffusion}, comprising $205$ samples with pre-defined drag points and masks. As is standard \cite{shi2023dragdiffusion,ling2023freedrag,cui2024stabledrag,liu2024drag}, we use the Image Fidelity (IF) and Mean Distance (MD) metrics for our analysis.
IF is calculated as $1-$LPIPS \cite{zhang2018unreasonable}, while MD assesses the accuracy with which handle points are moved to their designated targets. An ideal drag-based editing method would achieve a low MD, indicating effective drag editing, coupled with a high IF, signifying robust appearance preservation. 

\begin{figure*}
    \centering
    \includegraphics[width=0.925\textwidth]{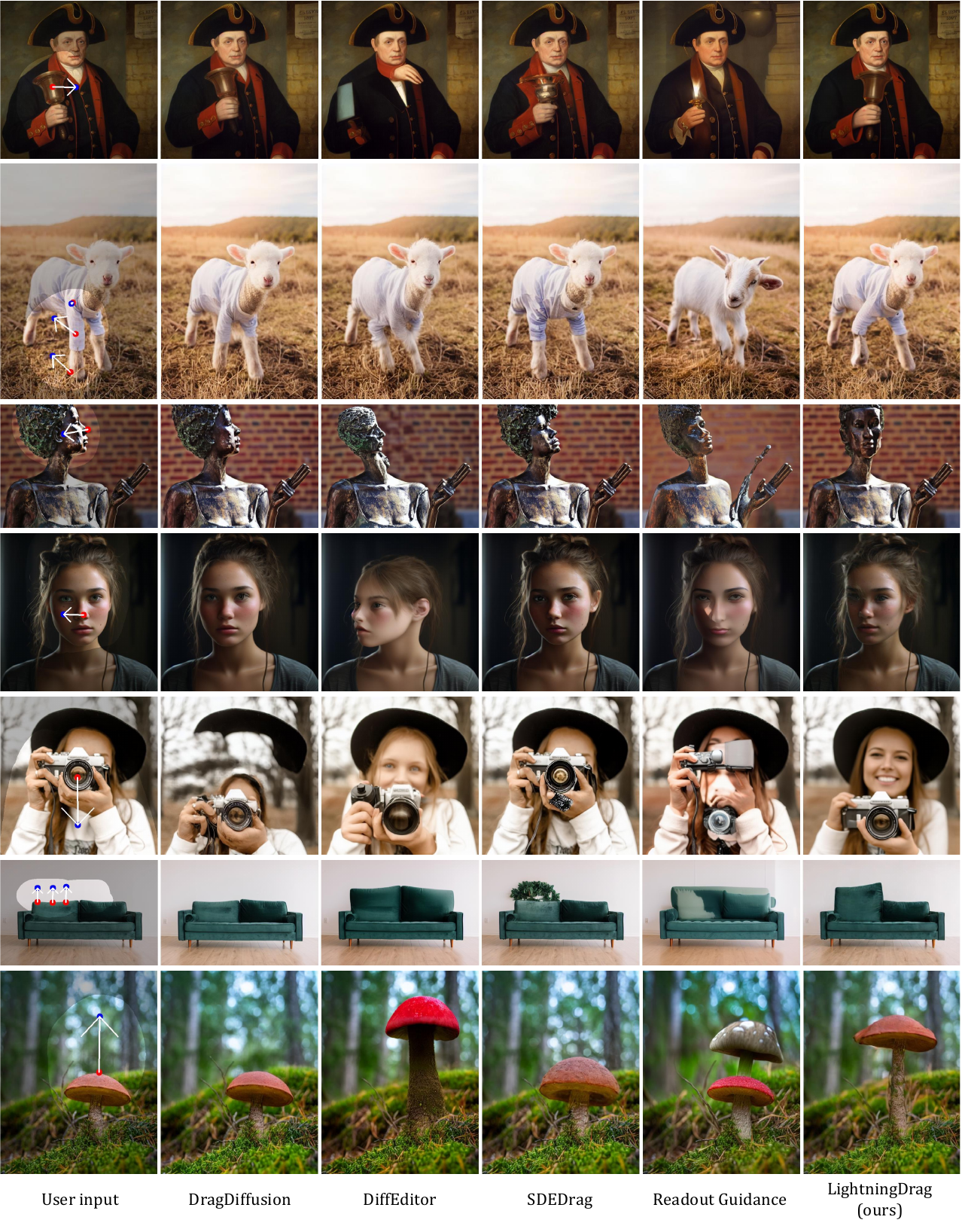}
    \vspace{-3mm}
    \caption{\textbf{Qualitative comparison on DragBench.} Our LightningDrag can handle various dragging instructions, such as pose change, scaling, translation \textit{etc.} while preserving the object identity.}
    \label{fig:qualitative_comparison}
\end{figure*}

\begin{figure*}
    \centering
    \includegraphics[width=0.83\textwidth]{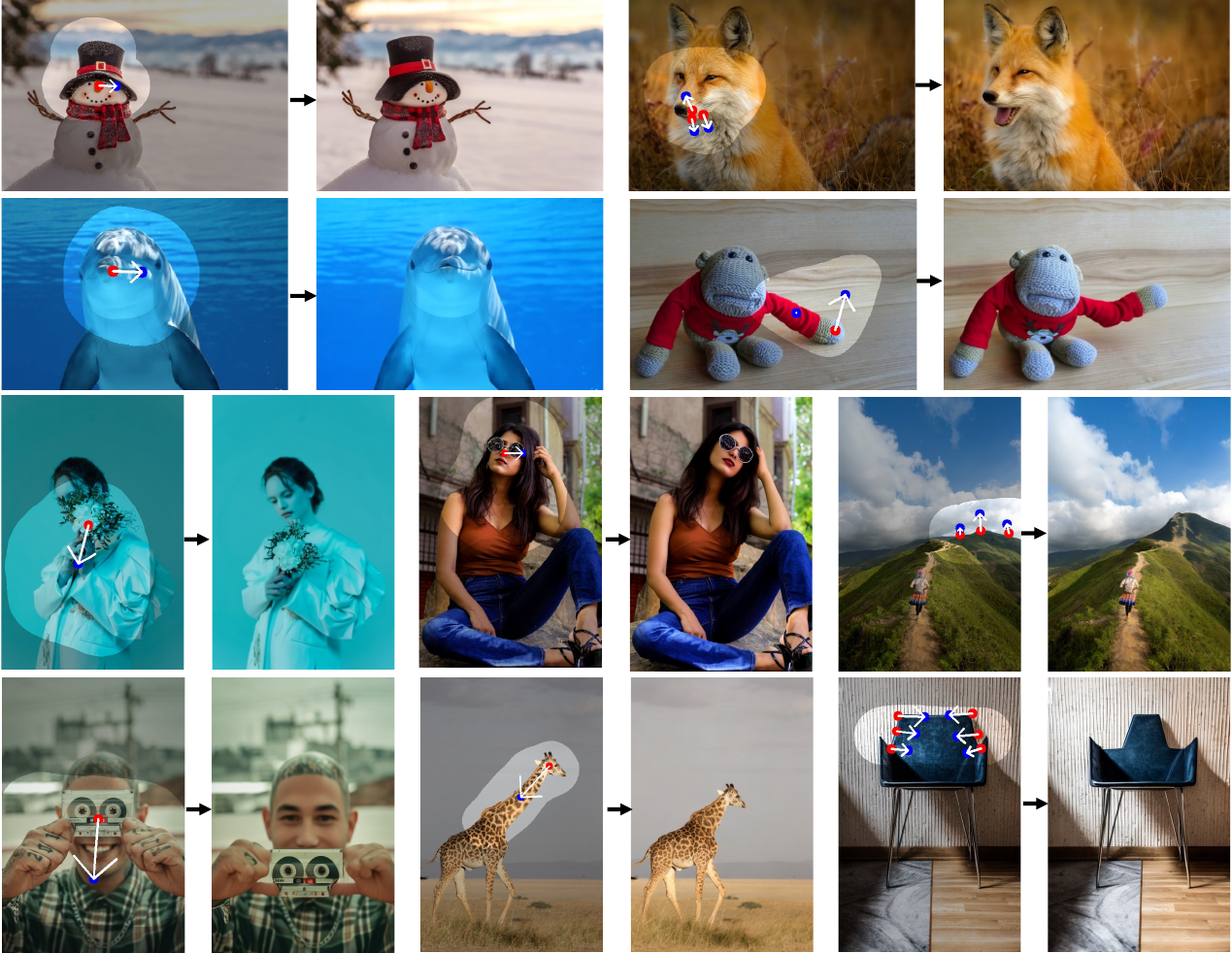}
    \vspace{-3mm}
    \caption{\textbf{Qualiative results of {\sc LightningDrag}.} Image credit (source images): Pexels.}
    \vspace{-1mm}
    \label{fig:more_qualitative}
\end{figure*}

Tab.~\ref{tab:quantitative} demonstrates the superiority of our {\sc LightningDrag} in term of point following, as evidenced by its lowest MD. We also notice that our {\sc LightningDrag} outperforms others in term of IF, except SDEDrag. However, further inspection reveals that SDEDrag often results in the undesired identity mapping (Fig. \ref{fig:qualitative_comparison} row 1, 3 and 4), leading to its high IF. Additional qualitative results supporting this observation are presented in Fig.~\ref{fig:more_qualitative}.

\begin{table}[]
    \centering
    \begin{tabular}{l c c}
         \hline
         Method & IF ($\uparrow$) & MD ($\downarrow$) \\
         \hline
         DragGAN \cite{pan2023drag} & $0.696$ & $57.155$ \\
         DragDiffusion \cite{shi2023dragdiffusion} & $0.881$ & $33.162$ \\
        DiffEditor \cite{mou2024diffeditor} & $0.856$ & $28.579$ \\
        Readout Guidance \cite{luo2023readout} & $0.790$ & $52.224$ \\
        SDEDrag \cite{nie2023blessing} & $\textbf{0.921}$ & $45.779$ \\
        \hline
        LightningDrag (ours) & $0.885$ & $\textbf{18.62}$ \\
        LightningDrag + LCM-LoRA (ours) & $0.890$ & $18.95$ \\
        \hline
    \end{tabular}
    \caption{\textbf{Quantitative comparison on DragBench.} IF and MD denote Image Fidelity (1-LPIPS) and Mean Distance, respectively.}
    \vspace{-5mm}
    \label{tab:quantitative}
\end{table}

\subsection{Time efficiency}
Due to the elimination of test-time latent optimization or gradient-based guidance, our {\sc LightningDrag} is extremely fast. Here, we compare the time efficiency of our {\sc LightningDrag} against the state-of-the-art methods. For fair comparisons, we extract a subset of square images from DragBench \cite{shi2023dragdiffusion}, resulting in $67$ images and perform inference at resolution of $512\times512$. We report the time cost on a NVIDIA A100 GPU. The results are shown in Tab.~\ref{tab:time_efficiency}. We notice that the execution time of SDEDrag has a high variance. This is because its inference speed depends on the distance between the handle and target points. In contrast, our {\sc LightningDrag} runs at a constant speed regardless of the dragging distance. Secondly, even without LCM-LoRA, our approach is already an order of magnitude faster than most baselines, making it suitable for practical applications. Lastly, when combined with recent diffusion acceleration methods such as LCM-LoRA, our LightningDrag can be further accelerated, requiring only $<1s$ for each dragging operation.

\begin{table}[]
    \centering
    \begin{tabular}{l c}
        \hline
        Model & Time (s)  \\
        \hline
        DragDiffusion \cite{shi2023dragdiffusion} & $55.05_{\pm 3.70}$ \\
        DiffEditor \cite{mou2024diffeditor} & $9.47_{\pm 0.20}$ \\
        Readout Guidance \cite{luo2023readout} & $98.13_{\pm 0.52}$ \\
        SDEDrag \cite{nie2023blessing} & $61.78_{\pm 18.84}$ \\
        \hline
        LightningDrag (ours) & $2.06_{\pm 0.05}$ \\
        LightningDrag + LCM-LoRA (ours) & $\textbf{0.92}_{\pm 0.02}$ \\
        \hline
        
    \end{tabular}
    \caption{\textbf{Time efficiency.} The reported time cost is obtained by running inference on 512$\times$512 images sampled from DragBench \cite{shi2023dragdiffusion} on a single NVIDIA A100 GPU.}
    \label{tab:time_efficiency}
\end{table}

\begin{figure*}
    \centering
    \includegraphics[width=0.83\textwidth]{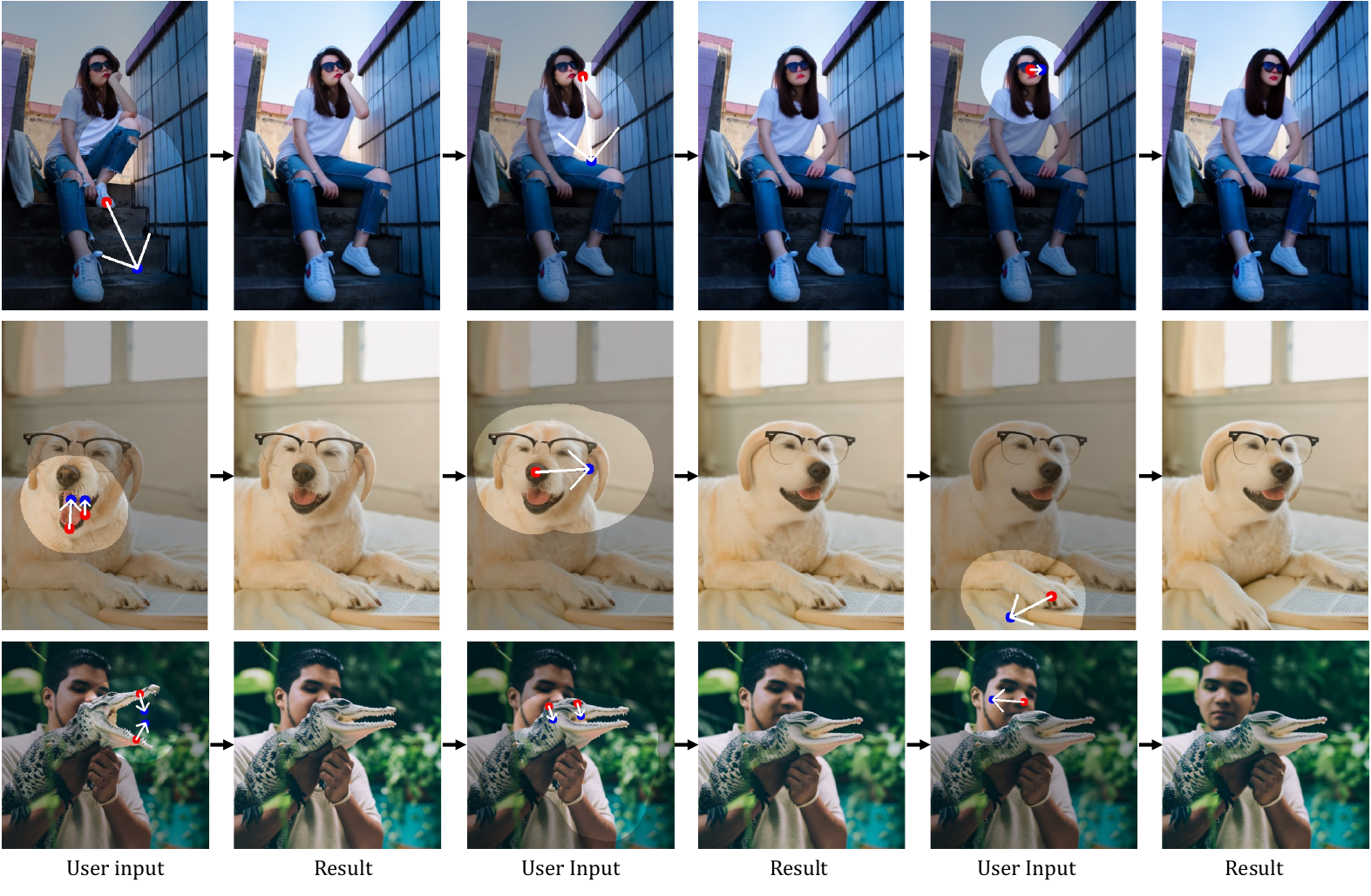}
    \vspace{-3mm}
    \caption{\textbf{Multi-round dragging.} 
    Image credit (source images): Pexels.}
    \label{fig:multi_round}
\end{figure*}

\subsection{Qualitative results}
\textbf{Comparisons with Prior Methods.}
We qualitatively compare our {\sc LightningDrag} with prior methods in Fig. \ref{fig:qualitative_comparison}. We observe that DiffEditor and Readout Guidance often struggle to preserve reference identity (\emph{e.g.}, 2nd and 3rd row), while DragDiffusion and SDEDrag sometimes fail to drag the regions-of-interest to the desired locations. In contrast, our {\sc LightningDrag} effectively handles various dragging needs, such as pose change, object scaling, translation, local deformation, while preserving the source image appearance.

\smallskip \noindent \textbf{Multi-round Dragging.}
Our {\sc LightningDrag} can be easily extended to multi-round dragging scenarios, where user can iteratively perform dragging based on the prior output. Examples of multi-round dragging are shown in Fig. \ref{fig:multi_round}.


\section{Conclusion, Limitations and Future Works}
We introduced {\sc LightningDrag}, a practical approach for high-quality drag-based image editing in $\sim 1s$. Despite the lack of training data, we show that natural video data contains rich motion cues, enabling the model to learn how objects change and deform. Extensive experiments demonstrate the superiority of our approach over prior methods in terms of both speed and quality. We hope our work can inspire future research on controllable and precise image editing.

However, since LightningDrag is built on Stable Diffusion V1.5, it inherits some of its limitations, such as inadequate detail in small regions, particularly with complex features like human hands and faces.
This limitation could potentially be mitigated by using larger diffusion models such as SDXL \cite{podell2023sdxl}, which we leave as future work.

\section*{Acknowledgement}
The authors would like to thank Zhongcong Xu, Zhijie Lin, Zilong Huang, Jianfeng Zhang for their helpful discussion and feedbacks.

{
    \small
    \bibliographystyle{ieeenat_fullname}
    \bibliography{main}
}


\end{document}